\documentclass{article}

\usepackage[preprint]{corl_2026} 

\usepackage[utf8]{inputenc} 
\usepackage[T1]{fontenc}    
\usepackage{hyperref}       
\usepackage{url}            
\usepackage{booktabs}       
\usepackage{amsfonts}       
\usepackage{nicefrac}       
\usepackage{microtype}      
\usepackage{xcolor}         

\usepackage{multirow}
\usepackage{amsmath}
\usepackage[table]{xcolor} 
\usepackage{graphicx}
\usepackage{wrapfig}
\usepackage{afterpage}
\usepackage{titlesec}
\usepackage{algorithm}
\usepackage{algpseudocode}
\usepackage{placeins}
\titlespacing*{\section}      {0pt}{1.0ex plus 0.2ex minus 0.1ex}{0.5ex plus 0.1ex}
\titlespacing*{\subsection}   {0pt}{0.7ex plus 0.2ex minus 0.1ex}{0.2ex plus 0.05ex}
\titlespacing*{\subsubsection}{0pt}{0.6ex plus 0.1ex minus 0.1ex}{0.2ex plus 0.05ex}
\AtBeginDocument{%
  \setlength{\abovedisplayskip}{4pt plus 1pt minus 1pt}%
  \setlength{\belowdisplayskip}{4pt plus 1pt minus 1pt}%
  \setlength{\abovedisplayshortskip}{2pt plus 1pt}%
  \setlength{\belowdisplayshortskip}{2pt plus 1pt minus 1pt}%
}
\definecolor{graybg}{HTML}{EFEFEF}
\definecolor{bluebg}{HTML}{E8E8FF}
\definecolor{greenbg}{HTML}{E8FFE8}

\title{StructRL: Structured Action-Space Exploration for Flow-Based VLAs}

\author{
  {\normalfont Jiarui Yang$^{1}$ \quad
  Bin Zhu$^{2}$ \quad
  Jingjing Chen$^{1,*}$ \quad
  Na Zou$^{3}$} \\
  {\normalfont Yanwei Fu$^{1}$ \quad
  Jianggang Zhu$^{1}$ \quad
  Yu-Gang Jiang$^{1}$} \\[0.4em]
  {\normalfont $^{1}$Fudan University \qquad
  $^{2}$Singapore Management University} \\
  {\normalfont $^{3}$Shanghai Artificial Intelligence Laboratory}
}

\begin{document}
\maketitle
\begingroup
\renewcommand{\thefootnote}{\fnsymbol{footnote}}
\footnotetext[1]{Corresponding author.}
\endgroup


\begin{abstract}
    Flow-based Vision-Language-Action (VLA) models are now widely used for continuous robotic manipulation, and online reinforcement learning (RL) is emerging as a key technique for adapting them to new tasks. Existing RL methods typically inject stochasticity inside the denoising chain, often through isotropic or temporally independent noise. However, effective robot exploration calls for structured noise: temporally smooth and scaled differently across action groups. We show that simply switching the in-chain noise to a structured form does not suffice: noise added at an intermediate flow time can be weakened by the remaining denoising steps before execution, a phenomenon we call \emph{Structured Noise Dilution}. We propose \textbf{StructRL}, which avoids dilution by relocating policy stochasticity to the action space via three coupled choices: (i) a deterministic ODE decoder, (ii) structured noise injected directly in the action space, and (iii) last-step replay, where policy-gradient updates avoid assigning likelihoods to intermediate denoising states. This keeps structured exploration tied to the executed action while providing a tractable training signal for the flow decoder. Across three flow-based VLA models on multiple simulated manipulation benchmarks and two real-world tasks, StructRL improves exploration efficiency and OOD performance over prior in-chain baselines, demonstrating the effectiveness of structured action-space exploration for adapting flow-based VLA with RL. \textbf{Project page:} \url{https://flyfaerss.github.io/structrl/}

\end{abstract}



\section{Introduction}

\begin{figure}[t]
    \centering
    \includegraphics[width=0.85\textwidth]{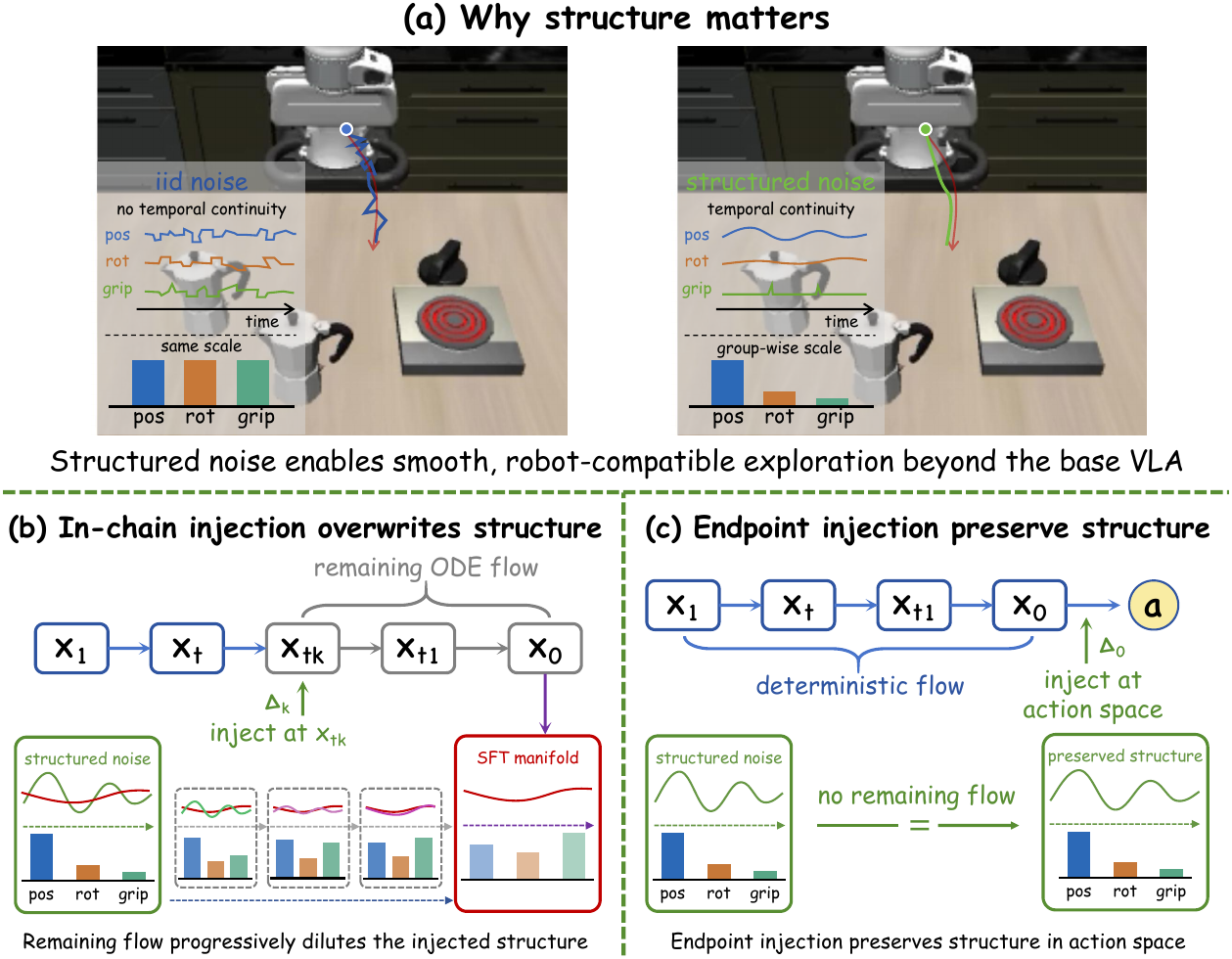}
    \caption{\textbf{Structured exploration for flow-based VLA-RL.}
    \textbf{(a)} Robot exploration benefits from smooth action chunks and group-aware noise scales.
    \textbf{(b)} In-chain noise can lose its intended structure during the remaining denoising steps.
    \textbf{(c)} StructRL keeps denoising deterministic and applies structured noise directly in the action space.}
    \label{fig:motivation}
\end{figure}

Vision-Language-Action (VLA) models cast robot control as instruction-conditioned action prediction, conditioning on visual observations and language instructions while typically pairing pretrained VLM backbones with continuous action heads trained on heterogeneous robot data~\citep{openvla, pi0, pi05, gr00t}. Among continuous action-head designs, flow matching~\citep{flowmatching} has become a common choice in recent VLA policies: it learns a velocity field and supports efficient few-step inference over action chunks. Recent flow-based VLAs such as $\pi_{0}$~\citep{pi0}, $\pi_{0.5}$~\citep{pi05}, and GR00T~\citep{gr00t} are trained by large-scale behavior cloning. This imitation objective, however, is fundamentally bounded by the quality and coverage of the offline demonstrations. Improving these pretrained policies beyond the offline data therefore requires feedback from environment interaction, motivating recent RL approaches for flow-based VLAs~\citep{pi_rl, stepnft}.

In these RL approaches, the exploration distribution determines which actions are evaluated during interaction and therefore shapes what the policy can learn. For continuous robot control, exploratory actions should reflect the structure of robot actions (Fig.~\ref{fig:motivation}a). First, actions sampled within a chunk should be temporally smooth, since jittery actions can make execution unstable or unsafe. Second, noise should be scaled differently across action groups, since \emph{position}, \emph{rotation}, and \emph{gripper} actions vary in range and safety sensitivity. Existing RL methods for flow-based VLAs~\citep{pi_rl,flow_cps,stepnft} instead inject random noise inside the denoising chain, without explicitly controlling temporal correlation or action-group scale at the action level.

We find that this in-chain design makes structured exploration difficult to realize (Fig.~\ref{fig:motivation}b). When structured noise is injected at an intermediate flow time, the remaining denoising steps map the perturbed sample back toward the SFT action manifold. As a result, the temporal and action-group structure specified by the injected noise can be weakened or overwritten before the action is executed. We refer to this phenomenon as \emph{Structured Noise Dilution} (Sec.~\ref{sec:dilution}).

This observation suggests a simple alternative: keep the flow decoder deterministic and place the exploration distribution directly in the action space (Fig.~\ref{fig:motivation}c). We propose \textbf{StructRL} based on this idea. Given the clean action chunk produced by ODE denoising, StructRL samples structured action-space noise from an AR(1)-correlated distribution with separate scales for action groups, conditioned on the decoder's terminal representation. During training, the policy likelihood is evaluated on the sampled action-space noise, so structured exploration stays tied to the executed action rather than to an intermediate denoising state. StructRL then uses last-step replay to pass policy-gradient updates to the flow decoder without defining likelihoods along the denoising chain. Across three flow-based VLA backbones ($\pi_{0}$, $\pi_{0.5}$, GR00T~N1.5) on multiple manipulation benchmarks, StructRL improves exploration efficiency and consistently outperforms prior in-chain baselines on out-of-distribution and long-horizon manipulation tasks. We further deploy $\pi_{0.5}$ on a Franka robot and instantiate the same action-space likelihood in an asynchronous AWAC~\citep{nair2020awac} learner. The comparisons between Gaussian exploration and StructRL on {\it pick-banana} and {\it plug-charger-in} tasks show faster and more effective online adaptation from sparse binary rewards. These results suggest that preserving structure in action-space exploration is important for efficient and robust VLA adaptation.

The main contributions can be summarized as follows:
\begin{itemize}\setlength\itemsep{0pt}\setlength\parsep{0pt}\setlength\topsep{2pt}
    \item We identify \emph{Structured Noise Dilution}: structured noise injected inside the flow chain can be weakened by the remaining denoising steps before it reaches the executed action.
    \item We introduce an action-space RL framework for flow-based VLA with last-step replay mechanism, showing that the flow decoder can be updated through the final Euler step alone, without assigning policy likelihoods to intermediate denoising states.
    \item Based on this framework, we propose \textbf{StructRL}, which injects structured exploration noise directly in the action space using AR(1) temporal correlation and separate scales for position, rotation, and gripper actions.
    \item Across multiple simulated manipulation benchmarks and two real-world tasks, we show that StructRL improves exploration efficiency and OOD performance over in-chain baselines and endpoint Gaussian exploration.
\end{itemize}

\section{Related Work}

\noindent\textbf{Vision-Language-Action Models.}
Early VLAs such as RT-1~\citep{rt1}, RT-2~\citep{rt2}, OpenVLA~\citep{openvla}, and $\pi_{0}$-FAST~\citep{pifast} represent robot actions as language tokens and train the model autoregressively. Another line keeps the vision--language backbone fixed and trains an attached continuous action head~\citep{octo, openvla_oft}. Recent VLA policies often use flow-matching~\citep{flowmatching, rectified_flow} or diffusion~\citep{diffusion_policy, chi_dp3} decoders to model continuous action chunks with few-step inference. Our work focuses on RL adaptation of representative flow-based VLAs, including $\pi_{0}$~\citep{pi0}, $\pi_{0.5}$~\citep{pi05}, and GR00T N1.5~\citep{gr00t}, which are trained by large-scale imitation learning on heterogeneous robot data.


\noindent\textbf{RL for Generative Robot Policies.}
RL has been used to optimize generative policies. DDPO~\citep{ddpo} treats the denoising process of a diffusion model as a multi-step stochastic policy and applies policy-gradient updates to the denoising chain. In offline continuous-control RL, Diffusion-QL~\citep{diffusion_ql}, IDQL~\citep{idql}, CPQL~\citep{cpql}, and QAM~\citep{qam} combine generative policy classes with value-based objectives. For flow-based VLAs, $\pi_{\text{RL}}$~\citep{pi_rl} introduces two methods: Flow-Noise models denoising as a discrete-time MDP with a learnable noise network, and Flow-SDE converts ODE denoising into an SDE for exploration. Flow-CPS~\citep{flow_cps} proposes a coefficient-preserving stochastic sampler for RL with flow matching, while $\pi$-StepNFT~\citep{stepnft} introduces a
critic and likelihood-free step-wise
negative-aware fine-tuning objective for
flow-based VLAs. These works make RL feasible for denoising policies, but their exploration signal is not defined as a structured distribution over executed actions.


\noindent\textbf{Structured Exploration in Continuous-Control RL.}
Continuous-control RL has explored several ways to structure exploration, including temporally correlated noise across environment steps (OU noise in DDPG~\citep{ou_noise}, colored noise~\citep{eberhard_pink, colored_ppo}), learned action-dimension coupling~\citep{lattice,yang2026actor}, state-dependent linear noise~\citep{gsde}, parameter-space perturbation~\citep{param_noise}, and on-manifold exploration~\citep{soe}. These methods mainly define structure on policy actions or parameters. Flow-based VLAs complicate this design because action samples are produced through a denoising chain, while existing RL methods usually place stochasticity inside the chain. StructRL moves the structured perturbation to the action space and uses last-step replay to pass policy-gradient updates back to the flow decoder.

\section{Structured Noise Dilution}
\label{sec:dilution}

This section studies why structure injected inside the denoising chain may not survive to the executed action. We first explain how the remaining denoising steps can filter an intermediate perturbation, then test this effect on a GR00T~N1.5 SFT model by injecting noise with controlled temporal correlation and per-group scale at different locations. We finally discuss why the structure induced by the SFT model is not a substitute for task-specific exploration.

\noindent\textbf{Mechanism.}\label{sec:dilution-theory}
In-chain exploration specifies noise before the decoder has finished predicting an action. The remaining denoising steps are still those of the SFT model, so the perturbed state is processed toward demonstration-like action regions. The action-level perturbation is therefore not the injected noise itself: its temporal correlation and action-group scale can be changed by the rest of the decoder, which we call \emph{Structured Noise Dilution}.

\noindent\textbf{Empirical Verification.}\label{sec:dilution-emp}
We test whether Structured Noise Dilution appears in flow-based VLAs, and Table~\ref{tab:dilution-summary} provides direct empirical evidence. Specifically, we use a GR00T~N1.5 SFT model on LIBERO-Long~\citep{libero}, and inject structured perturbations $\Delta$ with either a prescribed AR(1) coefficient $\rho_{\text{inj}}$ along the chunk axis or prescribed per-group magnitudes across position, rotation, and gripper actions. The perturbation is inserted at one of two sites: \emph{in-chain}, before the remaining denoising steps, or \emph{action-space}, after denoising completes. We then measure the structure of the resulting executed action.
The table reports one representative temporal-correlation setting and one representative action-group scale setting.

\newpage
\begin{wraptable}{r}{0.5\textwidth}
\vspace{-1.2em}
\centering
\footnotesize
\setlength\tabcolsep{2.5pt}
\renewcommand{\arraystretch}{1.0}
\caption{\textbf{Structured noise dilution.} Output statistics after injecting structured noise into a GR00T~N1.5 SFT model on LIBERO-Long. We compare action-space injection with in-chain injection under two requested structures: strong temporal correlation ($\rho_{\text{inj}}\!=\!0.95$) and a position-heavy scale ratio $R_{\text{approach}}\!=\!(1.0,0.5,0.05)$, both using base noise level $\sigma\!=\!0.2$. Action-space injection better follows the requested structure, while in-chain injection stays close to the SFT baseline.}
\label{tab:dilution-summary}
\begin{tabular}{l|ccc|ccc}
\toprule
 & \multicolumn{3}{c|}{\cellcolor{white}\textbf{Action-space}} & \multicolumn{3}{c}{\cellcolor{white}\textbf{In-chain}} \\
\cmidrule(lr){2-4}\cmidrule(lr){5-7}
 & pos & rot & grip & pos & rot & grip \\
\midrule
\rowcolor{gray!15} $\hat\rho_{\text{SFT}}$      & $0.78$ & $0.72$ & $0.57$ & $0.78$ & $0.72$ & $0.57$ \\
$\rho_{\text{inj}}\!=\!0.95$                    & $0.70$ & $0.64$ & $0.63$ & $0.79$ & $0.72$ & $0.58$ \\
\midrule
\rowcolor{gray!15} $\hat\sigma_{\text{SFT}}$    & $0.10$ & $0.08$ & $0.21$ & $0.10$ & $0.08$ & $0.21$ \\
$R_{\text{approach}}$                           & $0.23$ & $0.13$ & $0.19$ & $0.11$ & $0.08$ & $0.19$ \\
\bottomrule
\end{tabular}
\vspace{-1em}
\end{wraptable}

For strong temporal correlation ($\rho_{\text{inj}}\!=\!0.95$), action-space injection changes the measured output correlation, while in-chain injection stays close to the SFT baseline. For the position-heavy scale ratio $R_{\text{approach}}\!=\!(1.0,0.5,0.05)$ over (position, rotation, gripper), action-space injection increases the position standard deviation from $0.10$ to $0.23$, whereas in-chain injection changes it only from $0.10$ to $0.11$. Additional experimental details and full results are provided in App.~\ref{app:dilution-full}.

\noindent\textbf{SFT Structure Constrains
Exploration.}\label{sec:dilution-why}
The structure that survives in-chain injection comes from the SFT model. The SFT-induced correlation $\rho_{\text{SFT}}(o)$ reflects the behavior distribution in the offline data~\citep{ross2011dagger, atkeson1997robotlfd}, and may be useful as an imitation prior. However, if exploration is mainly shaped by this prior, the policy remains biased toward actions that the SFT model already favors. This makes it harder to reach useful actions outside the SFT prior and can reduce exploration efficiency during RL. These limitations motivate us to move structured exploration to the action space, where temporal correlation and action-group scale can be specified directly.

\section{Method: StructRL}
\label{sec:method}

\begin{wrapfigure}{r}{0.5\textwidth}
    \centering
    \vspace{-1.2em}
    \includegraphics[width=0.48\textwidth]{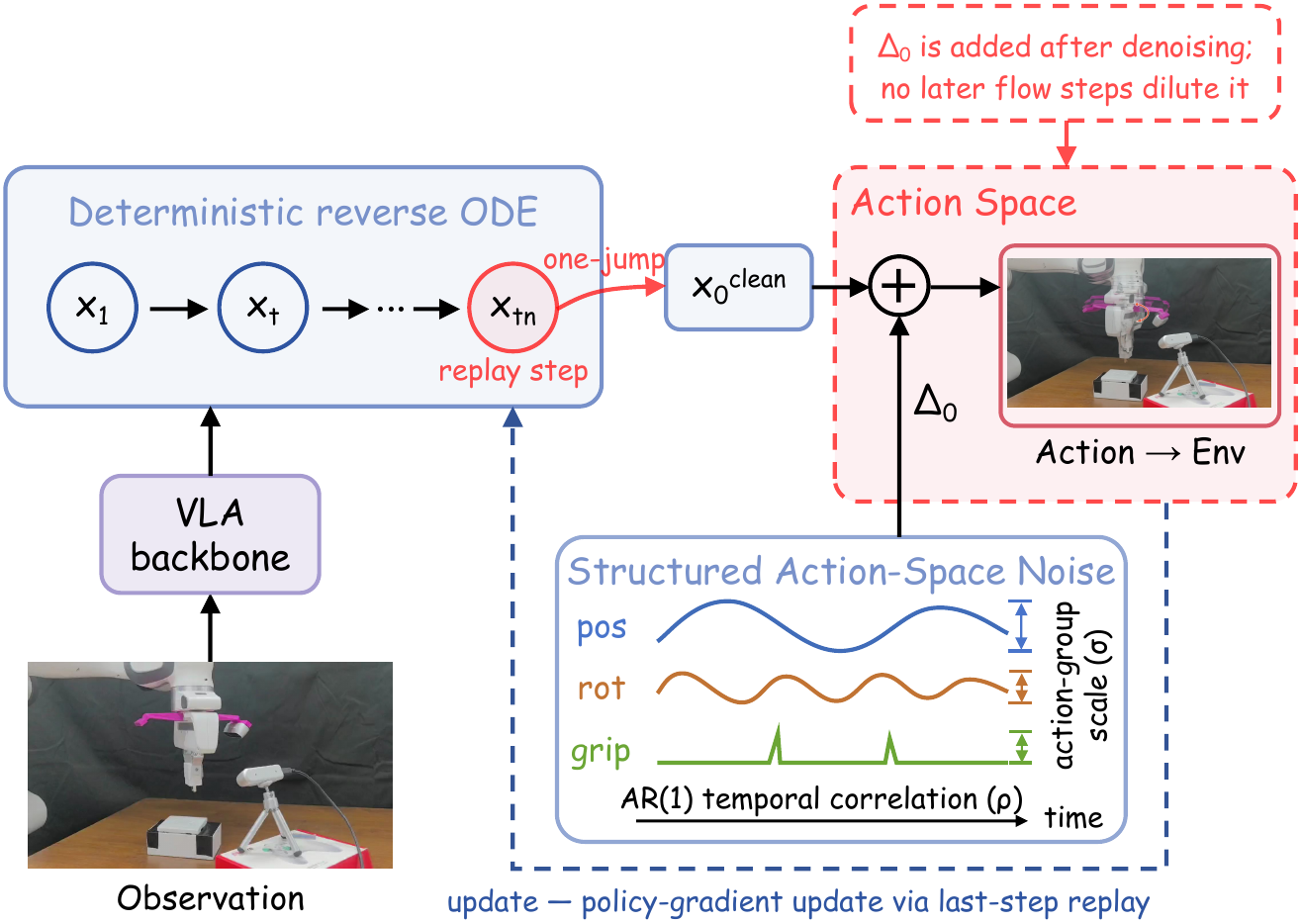}
    \caption{\textbf{The StructRL framework.} The reverse flow runs as a deterministic ODE to predict a clean actions $x_{0}^{\text{clean}}$. StructRL then samples structured noise $\Delta_{0}$ directly in the action space, with AR(1) temporal correlation and action-group-aware scales. During RL training, the likelihood is evaluated on this action-space noise, and last-step replay passes policy-gradient updates back to the flow decoder.}
    \label{fig:framework}
    \vspace{-1em}
\end{wrapfigure}
We propose StructRL, an action-space RL framework for flow-based VLAs, illustrated in Fig.~\ref{fig:framework}. StructRL keeps the flow decoder deterministic, injects structured noise at the action output, and uses last-step replay to update the decoder from the log-probability of the sampled action-space noise. In this section, we first review flow-based VLA policies and the RL likelihood problem, then describe the action-space RL framework, and finally instantiate the noise distribution with temporal correlation and action-group-aware scaling.

\subsection{Preliminaries}
\label{sec:prelim}

\noindent\textbf{Flow-Matching VLAs.}
A flow-based VLA predicts an action chunk by integrating a learned velocity field $v_{\theta}(x_t,t,o)$, where $o$ is the multimodal observation and $x_t\!\in\!\mathbb{R}^{C\times D}$ is a noised action chunk. Starting from $x_1\!\sim\!\mathcal{N}(0,I)$, the reverse flow is integrated with $T$ Euler steps:
\begin{equation}
x_{t_{k}-\delta} \;=\; x_{t_{k}} \;-\; \delta\, v_{\theta}(x_{t_{k}}, t_{k}, o), \qquad k = T, T\!-\!1, \ldots, 1,
\label{eq:euler}
\end{equation}
The final state $x_0$ is used as the clean action $a$. We denote by $F_{k\to0}^{\theta}$ the deterministic map from $x_{t_k}$ to $x_0$ obtained by the remaining $k$ denoising steps.

\noindent\textbf{Policy Likelihood for Flow-based RL.}
Likelihood-based RL objectives require a tractable policy density, such as policy ratios in PPO~\citep{ppo} or entropy terms in SAC~\citep{sac}. For a flow-matching VLA, the executed action is obtained after the multi-step denoising process in Eq.~\ref{eq:euler}, so its likelihood is not directly exposed. Existing flow-based VLA RL methods attach likelihoods to denoising-chain transitions~\citep{pi_rl}. Such chain-level likelihoods are tied to intermediate denoising states, while action-space structured noise is defined on the perturbation added to the executed output action.

\subsection{Action-Space RL with Last-Step Replay}
\label{sec:endpoint}
\label{sec:ppo}

Action-space exploration avoids dilution by the remaining denoising steps. Therefore, we first design a framework that can inject noise at the action output while still providing a tractable likelihood for RL. The pretrained flow decoder is run as a deterministic ODE to predict a clean action chunk,
\begin{equation}
x_{0}^{\text{clean}} \;=\; F_{T\to 0}^{\theta}(x_{1}),
\end{equation}
and then add noise at the action output:
\begin{equation}
a \;=\; x_{0}^{\text{clean}} \;+\; \Delta_{0}.
\label{eq:endpoint}
\end{equation}
Since $\Delta_{0}$ is added after denoising, the likelihood is evaluated on the perturbation applied to the executed action. As a simple example, let $\Delta_0 \sim \mathcal{N}(0,\sigma^2 I)$ in the action space. Its log-probability is
\begin{equation}
\log p_{\text{base}}(\Delta_{0})
\;=\;
\sum_{c,d}\log\mathcal{N}\!\left(\frac{\Delta_{0,c,d}}{\sigma};\,0,\,1\right)
\;-\;CD\log\sigma.
\label{eq:baseline-logprob}
\end{equation}

To update the flow decoder, we use last-step replay. During rollout, we store the executed action together with a stored late denoising state $x_{t_n}$. At each policy update, we replay the deterministic jump from $t_n$ to $0$:
\begin{equation}
\hat{x}_{0}^{\text{clean}} \;=\; F_{n\to 0}^{\theta}(x_{t_n}).
\label{eq:last-step-replay}
\end{equation}
In our default setting, $x_{t_n}$ is the last denoising state before the clean action chunk (i.e., $n=1$). During policy updates, this terminal jump is replayed to compute the log-probability without rerunning the full denoising chain. The action-space noise assigned to the stored action is then
\begin{equation}
\Delta_{0}^{\theta} \;=\; a - \hat{x}_{0}^{\text{clean}},
\label{eq:replay-noise}
\end{equation}
and the policy log-probability is evaluated as
\begin{equation}
\log \pi_{\theta}(a\mid o) \;=\; \log p_{\text{base}}(\Delta_{0}^{\theta}).
\label{eq:action-logprob}
\end{equation}
The likelihood is therefore evaluated for the executed action, rather than for an intermediate denoising-chain transition. This action-level log-probability can be used in either on-policy or off-policy likelihood-based objectives, with the value or critic loss defined by the chosen algorithm. In simulation, where rollouts can be collected in parallel, we use PPO. For real-world interaction, we use replay-based advantage-weighted actor--critic (AWAC)~\citep{awac} with an expectile value function~\citep{iql}, because physical rollouts are sequential and collected data must be reused many times.

Last-step replay updates the decoder through a short terminal denoising jump, while the policy likelihood is evaluated on the action-space noise. This avoids assigning likelihoods to intermediate denoising-chain transitions and provides the basic action-space RL framework used by both the Gaussian baseline and the structured noise in Sec.~\ref{sec:structure}. In Sec.~\ref{sec:ablations}, by setting $T\!\in\!\{2,3,4,5\}$, we demonstrate that this framework provides an effective optimization signal for the full flow field.

\subsection{Structured Noise Instantiation}
\label{sec:structure}

We now replace the Gaussian baseline in Sec.~\ref{sec:ppo} with a learnable structured distribution $p_{\phi}(\Delta_{0}\mid h_{\theta})$, where $h_{\theta}\!\in\!\mathbb{R}^{C\times d_{h}}$ is the decoder's final hidden representation. This distribution makes two properties explicit at the action output: temporal correlation within an action chunk and different noise scales across action groups.

\noindent\textbf{Action-Group-aware Scaling.}
Robot manipulation policies commonly represent actions as relative end-effector motion, typically covering position, rotation, and gripper control. These action groups have different units, ranges, and safety sensitivity, so a single isotropic noise scale is poorly matched to this action representation. We partition the $D$ action dimensions into three groups, $\mathcal{G} = \{\mathcal{G}_{\text{pos}}, \mathcal{G}_{\text{rot}}, \mathcal{G}_{\text{grip}}\}$. A global base noise level $\sigma^{\text{base}}$ sets the overall noise magnitude, while group-specific bounds control the allowed scale range for each action group. A lightweight MLP head $f_{\phi}$ predicts the bounded element-wise multipliers from $h_{\theta}$:
\begin{equation}
\sigma_{c,d} \;=\; \sigma^{\text{base}} \cdot \operatorname{clip}\!\big(f_{\phi}(h_{\theta})_{c,d},\, [\alpha_{g(d)},\, \beta_{g(d)}]\big),
\label{eq:group-scale}
\end{equation}
where $g(d)$ is the group of dimension $d$. The group-specific bounds $[\alpha_{g}, \beta_{g}]$ keep the learned scale within a controlled range.

\noindent\textbf{AR(1) Temporal Correlation.}
Within an action chunk, adjacent actions are executed in order, so independent noise across the chunk may introduce unnecessary jitter. Therefore, we further add temporal structure by sampling an AR(1) noise sequence. For each action dimension $d$,
\begin{equation}
\varepsilon_{1,d} \sim \mathcal{N}(0, 1), \qquad \varepsilon_{c+1,d} \;=\; \rho\,\varepsilon_{c,d} \;+\; \sqrt{1-\rho^{2}}\,\zeta_{c,d}, \quad \zeta_{c,d}\!\sim\!\mathcal{N}(0,1),
\label{eq:ar1}
\end{equation}
with $\rho\!\in\![0,1)$ controlling the temporal correlation. The final action-space perturbation combines the group-aware scale and the AR(1) noise sequence:
\begin{equation}
\Delta_{0,c,d} = \sigma_{c,d}\,\varepsilon_{c,d}.
\label{eq:structured-noise}
\end{equation}
This construction gives temporally correlated action-space noise with different scales across action groups, making the exploration distribution better aligned with the structure of robot actions.

\noindent\textbf{Structured log-probability.}
Given the replayed action-space perturbation $\Delta_{0}^{\theta}$, we evaluate its likelihood under the generative rule in Eq.~\ref{eq:structured-noise}. The AR(1) density is applied to the corresponding noise coordinates, with the scale change-of-variables term:
\begin{equation}
\resizebox{0.92\linewidth}{!}{$
\log p_{\phi}(\Delta_{0}^{\theta}\mid h_{\theta}) \;=\; \sum_{d=1}^{D}\!\left[\log\mathcal{N}(\varepsilon_{1,d}^{\theta};\,0,\,1) + \sum_{c=1}^{C-1}\log\mathcal{N}(\varepsilon_{c+1,d}^{\theta};\,\rho\varepsilon_{c,d}^{\theta},\,1\!-\!\rho^{2})\right] \;-\; \sum_{c,d}\log\sigma_{c,d},
$}
\label{eq:logprob}
\end{equation}
where $\varepsilon_{c,d}^{\theta}=\Delta_{0,c,d}^{\theta}/\sigma_{c,d}$ is used only for likelihood evaluation: it is the AR(1) noise value corresponding to the replayed perturbation under the current scale. This structured likelihood replaces Eq.~\ref{eq:baseline-logprob} inside the same action-space RL framework. We provide the detailed derivation in App.~\ref{app:logprob}.

\noindent\textbf{Regularization.}
The learned scale $\sigma$ controls action-space exploration, so unconstrained updates can collapse or over-amplify the
perturbations. Therefore, we regularize $\sigma$ with three lightweight terms: a dimension-entropy term that discourages scale collapse onto a few action dimensions, a budget term that keeps the overall noise magnitude near the target level, and a temporal smoothness term that avoids abrupt scale changes along the chunk. The full form is given in App.~\ref{app:regularizer}.

\begin{table}[t]
    \centering
    \small
    \setlength{\tabcolsep}{5pt}
    \renewcommand{\arraystretch}{1.05}
    \caption{\textbf{LIBERO benchmark results.} Average success rates (\%) across four task suites, each containing $10$ tasks. The few-shot SFT models are directly taken from the $\pi_{\text{RL}}$ checkpoints.}
    
    \label{tab:libero}
    \begin{tabular}{lcccccc@{\hskip 10pt}c}
    \toprule
    \textbf{Model} & \textbf{Spatial} & \textbf{Object} & \textbf{Goal} & \textbf{Long} & \textbf{Avg.} & \boldmath$\Delta$ \textbf{Avg.} \\
    \midrule
    \rowcolor{graybg}
    \textbf{GR00T N1.5} -- Few-shot SFT                  & 41.4 & 58.6 & 48.2 & 61.9 & 52.5 & --- \\
    \quad$+\,\pi_{\text{RL}}$ (Flow-SDE$+$PPO)  & 96.6 & \textbf{100}  & 93.8 & 95.6 & 96.5 & +44.0 \\
    \quad$+\,$Baseline                           & 96.2 & 99.4 & 91.8 & 95.6 & 95.8 & +43.3 \\
    \quad$+\,$StructRL                           & \textbf{99.2} & \textbf{100}  & \textbf{97.6} & \textbf{99.0} & \textbf{99.0} & \textbf{+46.3} \\
    \midrule
    \rowcolor{graybg}
    \boldmath$\pi_{0}$\unboldmath{} -- Few-shot SFT       & 65.3 & 64.4 & 49.8 & 51.2 & 57.6 & --- \\
    \quad$+\,\pi_{\text{RL}}$ (Flow-SDE$+$PPO)  & 98.4 & 99.4 & 96.2 & 90.2 & 96.0 & +38.4 \\
    \quad$+\,\pi$-StepNFT                        & 93.5 & 98.0 & 83.7 & 86.7 & 90.5 & +32.9 \\
    \quad$+\,$Baseline                           & \textbf{98.8} & 99.0 & 97.2 & 90.2 & 96.3 & +38.7 \\
    \quad$+\,$StructRL                           & \textbf{98.8} & \textbf{99.8} & \textbf{98.4} & \textbf{91.8} & \textbf{97.2} & \textbf{+39.6} \\
    \midrule
    \rowcolor{graybg}
    \boldmath$\pi_{0.5}$\unboldmath{} -- Few-shot SFT     & 84.6 & 95.4 & 84.6 & 43.9 & 77.1 & --- \\
    \quad$+\,\pi_{\text{RL}}$ (Flow-SDE$+$PPO)  & 99.6 & 100 & 98.8 & \textbf{93.0} & \textbf{97.9} & \textbf{+20.8} \\
    \quad$+\,\pi$-StepNFT                        & 97.8 & \textbf{100} & 98.2 & 79.8 & 94.0 & +16.9 \\
    \quad$+\,$Baseline                           & 99.4 & \textbf{100} & 93.8 & 86.0 & 94.8 & +17.7 \\
    \quad$+\,$StructRL                           & \textbf{99.6} & \textbf{100} & \textbf{98.8} & 90.2 & 97.2 & +20.1 \\
    \bottomrule
    \end{tabular}
    \end{table}

\section{Experiments}
\label{sec:experiments}
\subsection{Experimental Setup}
\noindent\textbf{Benchmarks.}
We conduct experiments on three standard simulation benchmarks: LIBERO~\citep{libero}, ManiSkill~\cite{maniskill}, and CALVIN~\citep{calvin}. Hyperparameter settings and CALVIN results are provided in App.~\ref{app:hparams} and App.~\ref{app:calvin}, respectively. All experiments are conducted on 4 NVIDIA A800 GPUs.

\noindent\textbf{Baseline variants.}
We adopt $\pi_{0}$~\citep{pi0}, $\pi_{0.5}$~\citep{pi05}, and GR00T~N1.5~\citep{gr00t} as the base VLA models throughout the experiments.
For simulation, we train the RL variants with PPO, since the environments allow parallel rollout collection. The physical experiments instead use the off-policy AWAC instantiation described in Sec.~\ref{sec:realrobot} and App.~\ref{app:realworld}.
StructRL is compared against several baselines: the SFT model, $\pi_{\text{RL}}$~\citep{pi_rl} (Flow-SDE$+$PPO), and $\pi$-StepNFT~\citep{stepnft}. Flow-SDE injects noise inside the denoising chain and trains with PPO, while $\pi$-StepNFT keeps SDE-based in-chain exploration but replaces likelihood-based policy-gradient training with a step-wise fine-tuning objective. We define the version that adds Gaussian noise within our action-space RL framework as \emph{Baseline}. StructRL further replaces this Gaussian noise with structured action-space noise.

\subsection{Main Results}
\noindent\textbf{Results on LIBERO.}
Table~\ref{tab:libero} presents the results across the four LIBERO suites. On GR00T~N1.5, where the few-shot SFT model reaches only 52.5 $\%$ average, StructRL pushes the average success rate to 99.0 $\%$, outperforming Flow-SDE$+$PPO by 2.5$\%$ with consistent gains on the four suites. On $\pi_{0}$, StructRL yields a modest improvement against Flow-SDE$+$PPO. Because all RL methods approach saturation on LIBERO, we treat this benchmark primarily as an in-distribution sanity check; ManiSkill OOD and the learning curves in Sec.~\ref{sec:ablations} provide the more diagnostic comparisons of injection location and structured exploration.

\noindent\textbf{Results on ManiSkill.}
Table~\ref{tab:ood} reports the simulated results on the ManiSkill benchmark. StructRL achieves the highest in-distribution success rate of 87.8$\%$ on $\pi_{0}$, and matches the best on $\pi_{0.5}$.
For OOD evaluation, the action-space RL framework shows a clear advantage over Flow-SDE's in-chain noise injection. Moving only the injection location from Flow-SDE to the endpoint Gaussian Baseline improves the OOD average by $28.0$ points on $\pi_{0}$ (39.3$\%\!\to\!67.3\%$) and $16.6$ points on $\pi_{0.5}$ (49.3$\%\!\to\!65.9\%$). This identifies endpoint action-space exploration, rather than AR structure alone, as the main source of the large final OOD gain. StructRL further changes the endpoint covariance and is most consistently beneficial in learning efficiency; its final values can be close to, or occasionally below, the Gaussian Baseline after saturation (e.g., the $\pi_0$ Execution split). Thus, the controlled results support two distinct conclusions: the injection location expands effective exploration beyond the flow-shaped SFT distribution, while temporal and group structure make that exploration more efficient.

\begin{table}[t]
\centering
\small
\setlength{\tabcolsep}{5pt}
\renewcommand{\arraystretch}{1.05}
\caption{\textbf{ManiSkill benchmark results.} Success rates (\%) on in-distribution (IND) tasks and three OOD splits: Vision (5 tasks), Semantic (5 tasks), and Execution (2 tasks). Avg. reports the mean success rate across the OOD splits.}
\label{tab:ood}
\begin{tabular}{lc@{\hskip 10pt}cccc}
\toprule
\multirow{2}{*}{\textbf{Model}} & \multirow{2}{*}{\textbf{IND}} & \multicolumn{4}{c}{\textbf{OOD}} \\ \cmidrule(l){3-6}
& & \textbf{Vision} & \textbf{Semantic} & \textbf{Execution} & \textbf{Avg.} \\
\midrule
\rowcolor{graybg}
\boldmath$\pi_{0}$\unboldmath{} -- SFT          & 38.4 & 32.6 & 8.4 & 13.2 & 18.1 \\
\quad$+\,\pi_{\text{RL}}$ (Flow-SDE$+$PPO)      & 78.8 & 61.1 & 25.4 & 31.5 & 39.3 \\
\quad$+\,\pi$-StepNFT                            & 79.2 & 69.1 & 49.1 & 33.1 & 50.4 \\
\quad$+\,$Baseline                               & 84.1 & 79.4 & 63.3 & \textbf{59.1} & \textbf{67.3} \\
\quad$+\,$StructRL                               & \textbf{87.8} & \textbf{80.0} & \textbf{64.2} & 56.2 & 66.8 \\
\midrule
\rowcolor{graybg}
\boldmath$\pi_{0.5}$\unboldmath{} -- SFT        & 40.1 & 38.8 & 16.6 & 22.3 & 25.9 \\
\quad$+\,\pi_{\text{RL}}$ (Flow-SDE$+$PPO)      & \textbf{90.9} & 68.0 & 34.5 & 45.4 & 49.3 \\
\quad$+\,\pi$-StepNFT                            & 85.4 & 76.9 & 56.6 & 45.1 & 59.5 \\
\quad$+\,$Baseline                               & 85.9 & 76.2 & 62.2 & 59.4 & 65.9 \\
\quad$+\,$StructRL                               & \textbf{90.9} & \textbf{78.3} & \textbf{68.4} & \textbf{64.2} & \textbf{70.3} \\
\bottomrule
\end{tabular}
\end{table}

\begin{figure}[!t]
    \centering
    \begin{minipage}[t]{0.24\textwidth}
        \centering
        \includegraphics[width=\linewidth]{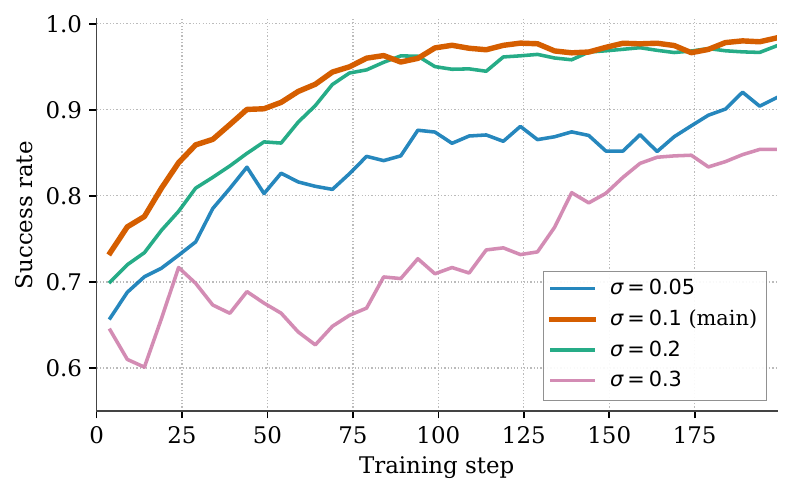}\\
        \scriptsize (a) Action-space noise level $\sigma^{\text{base}}$
    \end{minipage}
    \hfill
    \begin{minipage}[t]{0.24\textwidth}
        \centering
        \includegraphics[width=\linewidth]{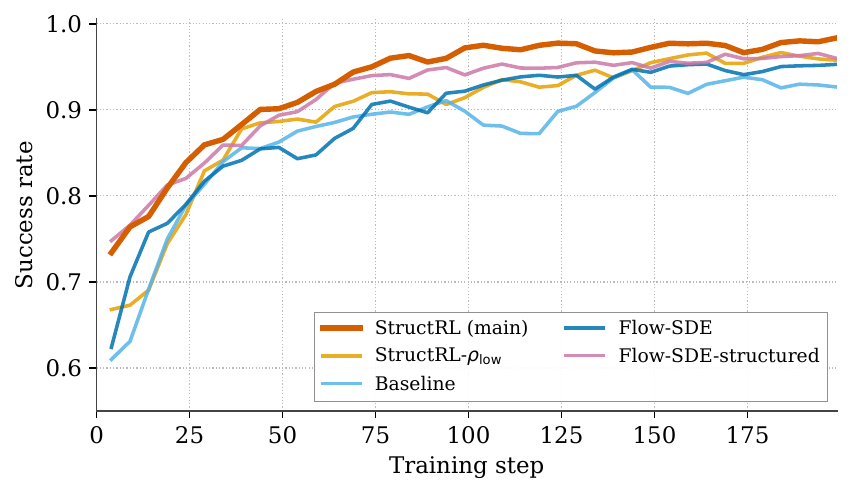}\\
        \scriptsize (b) Noise injection scheme
    \end{minipage}
    \hfill
    \begin{minipage}[t]{0.24\textwidth}
        \centering
        \includegraphics[width=\linewidth]{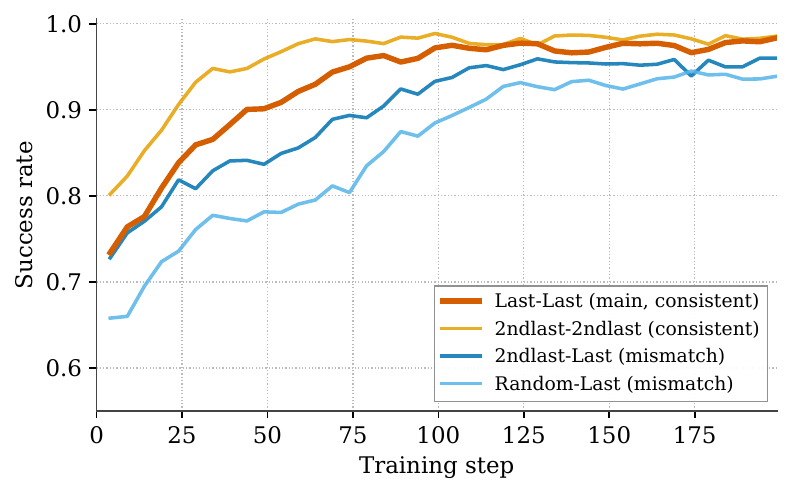}\\
        \scriptsize (c) Train--inference consistency
    \end{minipage}
    \hfill
    \begin{minipage}[t]{0.24\textwidth}
        \centering
        \includegraphics[width=\linewidth]{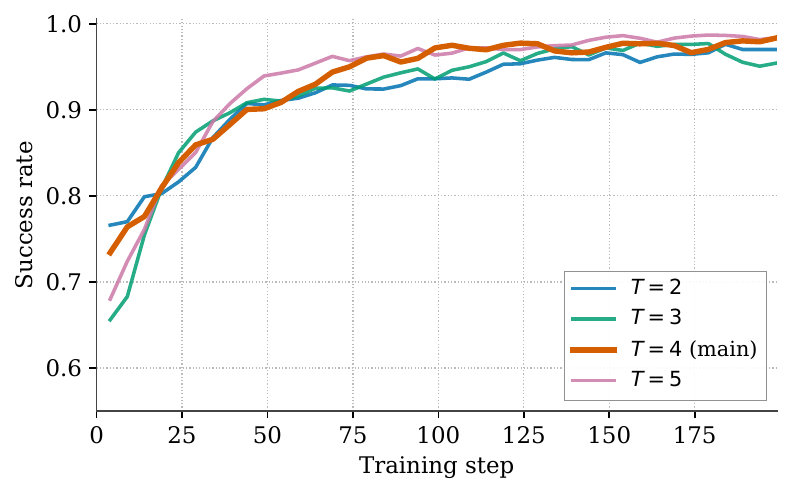}\\
        \scriptsize (d) Number of denoising steps $T$
    \end{minipage}
    \caption{\textbf{Four ablations on LIBERO-Long with GR00T~N1.5.} \textbf{(a)} Moderate action-space noise gives stable online improvement. \textbf{(b)} Structured action-space noise explores more efficiently than Gaussian or in-chain variants. \textbf{(c)} Matched train--inference replay improves performance, with larger matched jumps further improving exploration efficiency. \textbf{(d)} Last-step replay mechanism provides an effective optimization signal for the full flow field across different denoising steps.}
    \label{fig:ablation}
\end{figure}

\FloatBarrier
\subsection{Ablation Studies}
\label{sec:ablations}

All ablations use GR00T~N1.5 with PPO on LIBERO-Long, the LIBERO suite that most directly tests long-horizon execution. For a fair comparison, all runs use the same $200$ training iterations and are evaluated every $5$ iterations (Fig.~\ref{fig:ablation}). 

\noindent\textbf{(a) Sensitivity to the Noise Level $\sigma^{\text{base}}$.}
We set $\sigma^{\text{base}}\!\in\!\{0.05, 0.1, 0.2, 0.3\}$ to test how the magnitude of exploration noise affects training stability. As shown in Fig.~\ref{fig:ablation}(a), small noise levels under-explore, while large noise levels over-perturb the executed actions and destabilize training. The intermediate setting provides learnable signals over a broader action region, enabling efficient online improvement.

\noindent\textbf{(b) Noise Injection Scheme.}
We compare five ways of injecting exploration noise: \emph{action-space Gaussian noise} (the Baseline), \emph{StructRL}, \emph{StructRL$-$$\rho_{\text{low}}$} (reduced temporal correlation), \emph{Flow-SDE} (isotropic in-chain noise), and \emph{Flow-SDE$-$structured} (the same structured noise injected in-chain). These controls separate two factors. Flow-SDE versus the Gaussian Baseline changes the injection location, whereas the Gaussian Baseline versus StructRL changes only the endpoint noise structure. Fig.~\ref{fig:ablation}(b) shows that StructRL reaches 90$\%$ success in about 40 training steps, compared with about 80 steps for the endpoint Gaussian Baseline. The final values become close after saturation, so this ablation supports an efficiency benefit from temporal and group structure rather than a universal final-performance advantage.

\noindent\textbf{(c) Train--Inference Consistency.}
Last-step replay allows action-space RL to update the flow field by replaying a terminal ODE jump during training. This gives a flexible setting: the replay can cover one or more final denoising steps, and the inference-time ODE step can either match or differ from the training replay. We therefore compare four configurations. \textbf{Last-Last:} last-step replay ($n=1$) in training and last-step inference. \textbf{2ndlast-2ndlast:} second-last-step replay ($n=2$) in training and second-last-step inference. \textbf{2ndlast-Last:} second-last-step replay ($n=2$) in training but last-step inference. \textbf{Random-Last:} randomly selected replay depth in training and last-step inference. The first two settings are consistent between training and inference, while the last two deliberately introduce mismatch. As shown in Fig.~\ref{fig:ablation}(c), the train--inference consistent settings substantially outperform the mismatched settings. We also find that a larger replay jump brings a more evident gain in exploration efficiency.

\noindent\textbf{(d) Robustness Evaluation.}
To evaluate the robustness of last-step replay mechanism for our action-space RL framework, we vary the number of denoising steps as $T\!\in\!\{2,3,4,5\}$. As shown in Fig.~\ref{fig:ablation}(d), all settings show similar convergence speeds and achieve comparable final performance. \textbf{This result shows that last-step replay can route gradients only through a terminal denoising jump while still providing an efficient optimization signal for the full flow field.}

\FloatBarrier
\subsection{Real-World Experiment}
\label{sec:realrobot}

\begin{figure}[t]
    \centering
    \begin{minipage}[c]{0.42\linewidth}
        \centering
        \begin{minipage}[c]{0.49\linewidth}
            \centering
            \includegraphics[width=\linewidth]{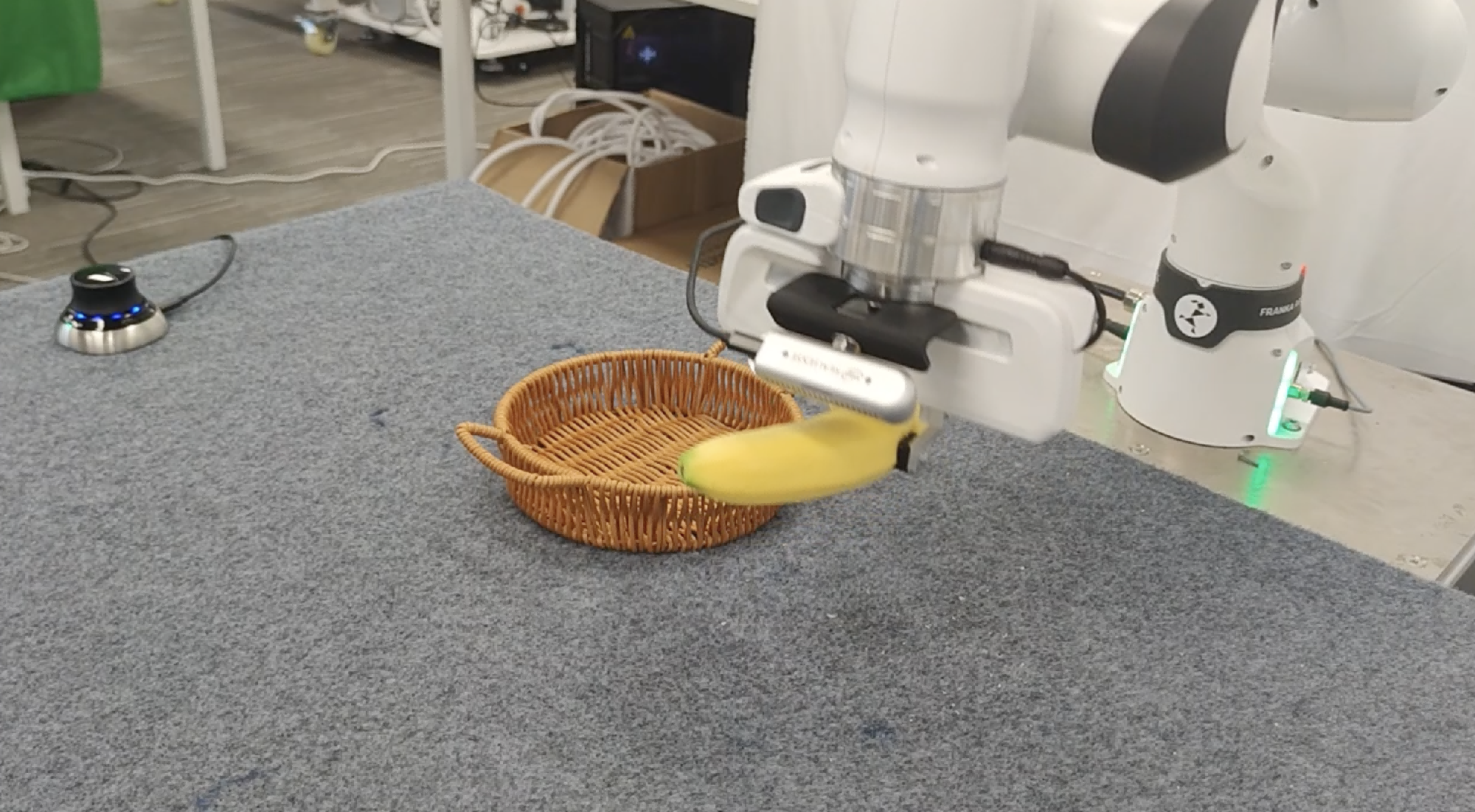}\\[-0.2em]
            \scriptsize Pick-Banana
        \end{minipage}
        \hfill
        \begin{minipage}[c]{0.49\linewidth}
            \centering
            \includegraphics[width=\linewidth]{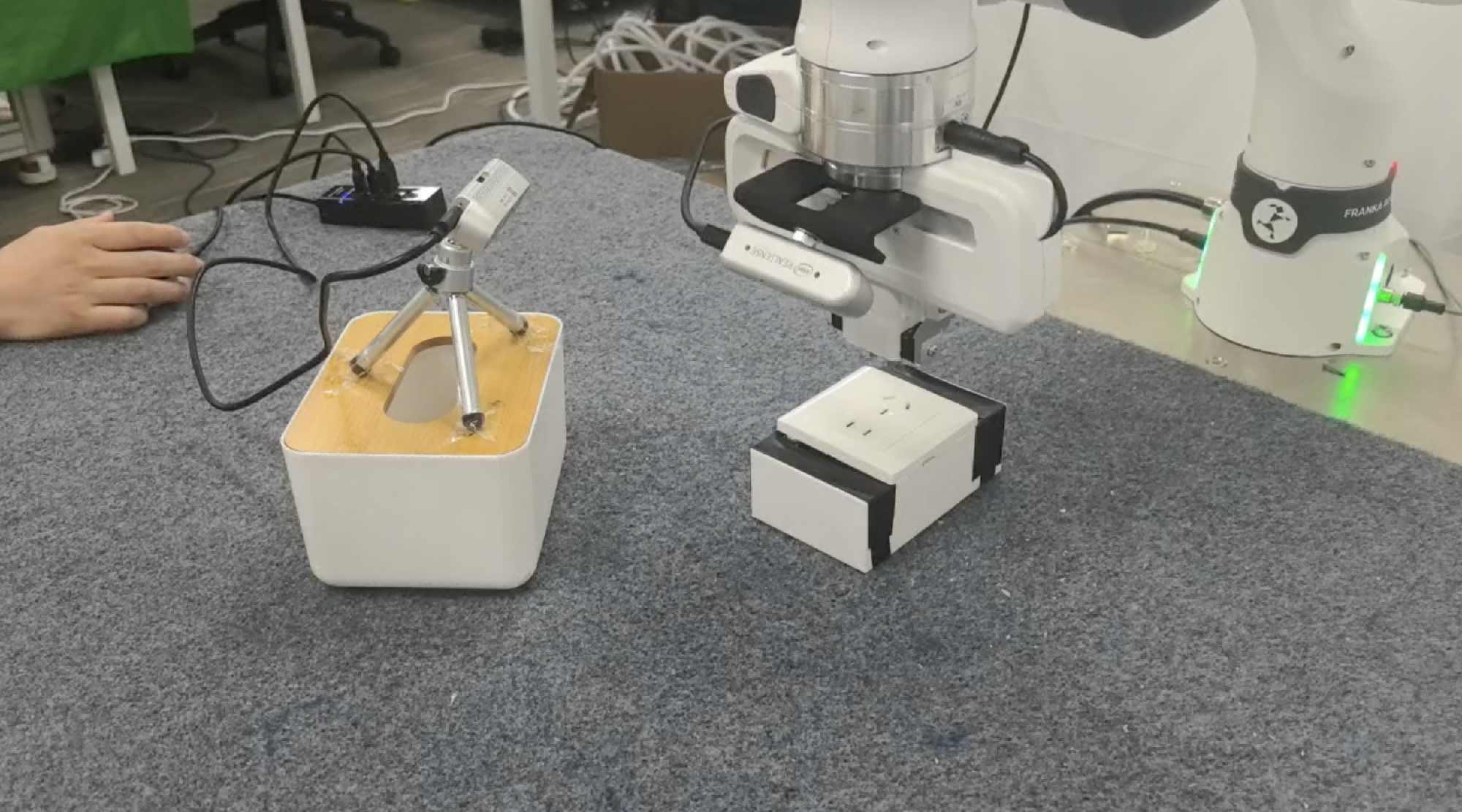}\\[-0.2em]
            \scriptsize Plug-Charger-In
        \end{minipage}
    \end{minipage}
    \hfill
    \begin{minipage}[c]{0.55\linewidth}
        \centering
        {\small
        \setlength{\tabcolsep}{3.5pt}
        \renewcommand{\arraystretch}{1.08}
        \begin{tabular}{lcc}
        \toprule
        \textbf{Task / metric} & \textbf{Baseline} & \textbf{StructRL} \\
        \midrule
        Pick-banana, SR at 60 min & 56\% (14/25) & \textbf{84\% (21/25)} \\
        Plug-charger-in, convergence & $\sim$35 min & \textbf{$\sim$30 min} \\
        \bottomrule
        \end{tabular}}
    \end{minipage}
    \caption{\textbf{Real-world tasks and online-RL comparisons.} Both task-specific policies are initialized by SFT on 10 pre-collected demos and achieve 0\% success rate (SR) before online RL.}
    \label{fig:realrobot}
    \vspace{-0.8em}
\end{figure}

We evaluate StructRL in the real world on \emph{pick-banana}, a pick-and-place (PnP) task, and \emph{plug-charger-in}, a fine-manipulation task (Fig.~\ref{fig:realrobot}). Because physical rollouts have low throughput and each transition is expensive to collect, we combine StructRL with AWAC~\cite{nair2020awac} to obtain an off-policy real-world RL algorithm that can repeatedly reuse collected data.

\noindent\textbf{Algorithm:}
To retain the unlimited transition-reuse advantage of off-policy learning, we set the ODE denoising horizon to a single step ($T=1$). The latent saved by last-step replay is therefore the initial Gaussian noise $x_{1}$, which is independent of the policy that collected the transition. Each chunk-based transition is stored as
\begin{equation}
\mathcal{T}_t=\big(o_t,a_t^b,\mathbf{r}_t,\mathbf{d}_t,o_{t+C},x_{1}\big),
\label{eq:real_transition}
\end{equation}
where $o_t$ and $o_{t+C}$ are the current and next observations; $a_t^b$ is the executed action chunk, either generated by the policy with exploration noise or human intervention; $\mathbf{r}_t$ and $\mathbf{d}_t$ are the rewards and termination/truncation indicators within the chunk. Given the stored $x_{1}$, the current actor recomputes the one-step action chunk without noise injection and evaluates $a_t^b$ under the StructRL action-space distribution. Its AWAC objective is
\begin{equation}
\mathcal{L}_{\mathrm{AWAC}}(\theta)
=-\mathbb{E}_{\mathcal{T}_t\sim\mathcal{D}}
\left[w_t\log\pi_\theta(a_t^b\mid o_t,x_{1})\right],
\label{eq:real_awac}
\end{equation}
where the advantage weight is
\begin{equation}
w_t=\exp\!\left(\operatorname{clip}\!\left(
\frac{\min_i Q'_{i}(o_t,a_t^b)-V(o_t)}{\beta},-3,3\right)\right),
\qquad \beta=0.5.
\label{eq:real_awac_weight}
\end{equation}
Thus, previously collected rollout actions and human corrections can be repeatedly reused through advantage-weighted BC, while higher-advantage actions receive larger likelihood updates.

\noindent\textbf{Setup:}
We conduct the experiments on a FR3 robot. {\it Pick-banana} uses $C=10$ and a 7-DoF action while {\it Plug-charger-in} uses $C=5$ and a 6-DoF action with fixed gripper dimension. Both tasks use delta end-effector pose actions to represent action space.

\noindent\textbf{Results:}
For each task, we first perform SFT on ten pre-collected demos. This initialization provides only a coarse motion trend and achieves 0\% success rate. We then run online RL with few human interventions. In {\it pick-banana}, the limited demos do not cover the banana's varying positions, so the initial policy usually fails during grasping. After 60 minutes of online RL, StructRL reaches an 84\% success rate (21/25), whereas the Gaussian baseline reaches 56\% (14/25), as shown in Fig.~\ref{fig:realrobot}. For {\it plug-charger-in}, the SFT prior mainly produces a downward insertion motion but cannot make the necessary lateral alignment corrections. StructRL reaches a stable, fully successful regime after approximately 30 minutes, while the Gaussian baseline requires approximately 35 minutes (Fig.~\ref{fig:realrobot}). These results show that structured exploration also improves exploration efficiency in real-world online RL. Additional implementation, training, and safety details are provided in App.~\ref{app:realworld}.

\FloatBarrier

\section{Conclusion}
\label{sec:conclusion_main}

We presented StructRL, an action-space RL framework for RL adaptation of flow-based VLAs. Motivated by structured noise dilution, StructRL keeps the flow decoder deterministic, injects temporally correlated and group-aware noise on the executed action chunk, and uses last-step replay to update the flow field from action-space likelihoods. Controlled simulation results show that moving exploration to the action space accounts for most of the OOD gain, while structuring the action-space noise primarily improves learning efficiency. Real-world experiments further show that StructRL enables more efficient online adaptation than Gaussian exploration.

\bibliography{example_new}  

\appendix
\renewcommand{\thesection}{\Alph{section}}
\setcounter{section}{0}
\setcounter{table}{0}
\setcounter{figure}{0}
\makeatletter
\@addtoreset{table}{section}
\@addtoreset{figure}{section}
\makeatother
\renewcommand{\thetable}{\Alph{section}\arabic{table}}
\renewcommand{\thefigure}{\Alph{section}\arabic{figure}}

\section{StructRL Training Pseudocode}
\label{app:pseudocode}

This appendix provides the training pseudocode for the framework introduced in Sec.~\ref{sec:endpoint}. Algorithm~\ref{alg:structrl} summarizes the action-space RL framework used by both the Gaussian baseline and StructRL. During rollout, the flow decoder first predicts a clean action through the deterministic ODE, and exploration noise is then added in the action space. During policy updates, last-step replay recomputes the clean action from a stored late denoising state, so the likelihood can be evaluated on the action-space perturbation assigned to the executed action.

\begin{algorithm}[h]
\caption{StructRL}
\label{alg:structrl}
\begin{algorithmic}[1]
\Require Flow decoder $F^{\theta}$ initialized from the SFT model; action-space noise distribution $p$; likelihood-based RL objective $\mathcal{L}_{\text{RL}}$
\For{each rollout iteration}
  \State Sample $x_{1}\!\sim\!\mathcal{N}(0,I)$ and run the flow decoder as a deterministic ODE.
  \State Store a late denoising state $x_{t_n}$ and obtain $x_{0}^{\text{clean}}=F_{T\to0}^{\theta}(x_1)$.
  \State Sample action-space noise $\Delta_0\!\sim\!p$, where $p$ is either the Gaussian baseline or the structured distribution defined by Eq.~\ref{eq:structured-noise}.
  \State Execute $a=x_{0}^{\text{clean}}+\Delta_0$ and store the transition, including $a$, $x_{t_n}$, and the rollout log-probability when required by the RL objective.
\EndFor
\For{each policy update}
  \State Replay the terminal denoising jump from the stored state: $\hat{x}_{0}^{\text{clean}}=F_{n\to0}^{\theta}(x_{t_n})$.
  \State Recover the replayed action-space perturbation $\Delta_{0}^{\theta}=a-\hat{x}_{0}^{\text{clean}}$.
  \State Evaluate the action-level log-probability using Eq.~\ref{eq:baseline-logprob} for the Gaussian baseline or Eq.~\ref{eq:logprob} for StructRL.
  \State Optimize $\mathcal{L}_{\text{RL}}$ with this action-level log-probability.
\EndFor
\end{algorithmic}
\end{algorithm}

\begin{table}[!htbp]
    \centering
    \small
    \setlength\tabcolsep{4pt}
    \renewcommand{\arraystretch}{1.0}
    \caption{\textbf{Full per-condition data at base noise level $\sigma\!=\!0.2$.} GR00T~N1.5 SFT model on LIBERO-Long. \textbf{(a)} Lag-1 autocorrelation $\hat\rho_{\text{out}}$ under different injected $\rho_{\text{inj}}$. \textbf{(b)} Per-group standard deviation $\hat\sigma_{\text{out}}$ under different magnitude profiles (target $\sigma_{\text{inj}}\!=\!0.2\cdot R$). Action-space injection tracks the requested structure, while in-chain injection stays close to the SFT baseline.}
    \label{tab:dilution}
    \begin{tabular}{l|ccc|ccc}
    \toprule
    \multirow{2}{*}{\textbf{Injected structure}} & \multicolumn{3}{c|}{\textbf{Action-space output}} & \multicolumn{3}{c}{\textbf{In-chain output}} \\
    \cmidrule(lr){2-4}\cmidrule(lr){5-7}
     & pos & rot & grip & pos & rot & grip \\
    \midrule
    \multicolumn{7}{l}{\emph{(a) Lag-1 autocorrelation $\hat\rho_{\text{out}}$}} \\
    \rowcolor{gray!15} SFT baseline ($\rho_{\text{SFT}}$, no injection)             & $0.784$ & $0.718$ & $0.566$ & $0.784$ & $0.718$ & $0.566$ \\
    $\rho_{\text{inj}}\!=\!0.00$ \ (no banding)                                      & $0.115$ & $-0.033$ & $0.225$ & $0.785$ & $0.718$ & $0.564$ \\
    $\rho_{\text{inj}}\!=\!0.30$                                                     & $0.310$ & $0.203$ & $0.378$ & $0.787$ & $0.721$ & $0.562$ \\
    $\rho_{\text{inj}}\!=\!0.70$                                                     & $0.556$ & $0.493$ & $0.560$ & $0.788$ & $0.725$ & $0.577$ \\
    $\rho_{\text{inj}}\!=\!0.95$ \ (strong banding)                                  & $0.705$ & $0.643$ & $0.634$ & $0.789$ & $0.723$ & $0.577$ \\
    \textbf{output range} ($\max\!-\!\min$ over $\rho_{\text{inj}}\!\in\![0,0.95]$)  & $0.590$ & $0.676$ & $0.409$ & $0.004$ & $0.007$ & $0.015$ \\
    \midrule
    \multicolumn{7}{l}{\emph{(b) Per-group standard deviation $\hat\sigma_{\text{out}}$ (target $\sigma_{\text{inj}}\!=\!0.2\cdot R$)}} \\
    \rowcolor{gray!15} SFT baseline ($\sigma_{\text{SFT}}$, no injection)            & $0.105$ & $0.077$ & $0.206$ & $0.105$ & $0.077$ & $0.206$ \\
    $R_{\text{approach}}\!=\!(1.0,0.5,0.05)$ \ (pos-heavy)                            & $0.227$ & $0.128$ & $0.192$ & $0.109$ & $0.078$ & $0.193$ \\
    $R_{\text{pregrasp}}\!=\!(0.05,0.1,1.0)$ \ (grip-heavy)                           & $0.106$ & $0.080$ & $0.314$ & $0.106$ & $0.077$ & $0.214$ \\
    $R_{\text{carry}}\!=\!(0.5,0.5,0.1)$ \ (pos$+$rot)                                & $0.147$ & $0.127$ & $0.205$ & $0.106$ & $0.077$ & $0.190$ \\
    \textbf{output range} ($\max\!-\!\min$ across the three ratios)                  & $0.121$ & $0.048$ & $0.122$ & $0.003$ & $0.001$ & $0.024$ \\
    \bottomrule
    \end{tabular}
    
    \vspace{0.8em}
    
    \caption{\textbf{Full per-condition data at base noise level $\sigma\!=\!0.5$.} Same GR00T~N1.5 SFT model on LIBERO-Long as Table~\ref{tab:dilution}, with a larger base noise level (target $\sigma_{\text{inj}}\!=\!0.5\cdot R$). The same dilution pattern appears: action-space injection follows the requested structure, while in-chain injection remains close to the SFT baseline.}
    \label{tab:dilution-sigma05}
    \begin{tabular}{l|ccc|ccc}
    \toprule
    \multirow{2}{*}{\textbf{Injected structure}} & \multicolumn{3}{c|}{\textbf{Action-space output}} & \multicolumn{3}{c}{\textbf{In-chain output}} \\
    \cmidrule(lr){2-4}\cmidrule(lr){5-7}
     & pos & rot & grip & pos & rot & grip \\
    \midrule
    \multicolumn{7}{l}{\emph{(a) Lag-1 autocorrelation $\hat\rho_{\text{out}}$}} \\
    \rowcolor{gray!15} SFT baseline ($\rho_{\text{SFT}}$, no injection)             & $0.780$ & $0.723$ & $0.584$ & $0.780$ & $0.723$ & $0.584$ \\
    $\rho_{\text{inj}}\!=\!0.00$ \ (no banding)                                      & $-0.038$ & $-0.061$ & $0.275$ & $0.785$ & $0.737$ & $0.589$ \\
    $\rho_{\text{inj}}\!=\!0.30$                                                     & $0.192$ & $0.181$ & $0.390$ & $0.797$ & $0.752$ & $0.597$ \\
    $\rho_{\text{inj}}\!=\!0.70$                                                     & $0.489$ & $0.484$ & $0.558$ & $0.812$ & $0.772$ & $0.612$ \\
    $\rho_{\text{inj}}\!=\!0.95$ \ (strong banding)                                  & $0.644$ & $0.635$ & $0.609$ & $0.803$ & $0.759$ & $0.617$ \\
    \textbf{output range} ($\max\!-\!\min$ over $\rho_{\text{inj}}\!\in\![0,0.95]$)  & $0.682$ & $0.696$ & $0.334$ & $0.027$ & $0.035$ & $0.028$ \\
    \midrule
    \multicolumn{7}{l}{\emph{(b) Per-group standard deviation $\hat\sigma_{\text{out}}$ (target $\sigma_{\text{inj}}\!=\!0.5\cdot R$)}} \\
    \rowcolor{gray!15} SFT baseline ($\sigma_{\text{SFT}}$, no injection)            & $0.105$ & $0.077$ & $0.208$ & $0.105$ & $0.077$ & $0.208$ \\
    $R_{\text{approach}}\!=\!(1.0,0.5,0.05)$ \ (pos-heavy)                            & $0.511$ & $0.262$ & $0.209$ & $0.116$ & $0.080$ & $0.199$ \\
    $R_{\text{pregrasp}}\!=\!(0.05,0.1,1.0)$ \ (grip-heavy)                           & $0.109$ & $0.093$ & $0.572$ & $0.107$ & $0.077$ & $0.249$ \\
    $R_{\text{carry}}\!=\!(0.5,0.5,0.1)$ \ (pos$+$rot)                                & $0.273$ & $0.262$ & $0.206$ & $0.109$ & $0.081$ & $0.197$ \\
    \textbf{output range} ($\max\!-\!\min$ across the three ratios)                  & $0.402$ & $0.169$ & $0.366$ & $0.009$ & $0.004$ & $0.052$ \\
    \bottomrule
    \end{tabular}
    \end{table}

\section{Full Empirical Verification of Structured Noise Dilution}
\label{app:dilution-full}

This appendix expands the empirical verification in Sec.~\ref{sec:dilution-emp}. Table~\ref{tab:dilution-summary} in the main paper reports one representative temporal-correlation setting and one representative action-group scale setting. Here we provide the full sweep at $\sigma\!=\!0.2$ (Table~\ref{tab:dilution}) and an additional run at $\sigma\!=\!0.5$ (Table~\ref{tab:dilution-sigma05}). The larger noise level checks that the observed dilution is not an artifact of using a small perturbation.

\noindent\textbf{Setup.}
We use a GR00T~N1.5 SFT model on LIBERO-Long and inject structured perturbations $\Delta$ in two ways. For temporal structure, $\Delta$ is sampled with a prescribed AR(1) coefficient $\rho_{\text{inj}}$ along the chunk axis. For action-group scale, $\Delta$ uses target magnitudes $\sigma_{\text{inj}}\!=\!\sigma\cdot R$ over the (position, rotation, gripper) groups, where $\sigma$ is the base noise level and $R$ is a ratio vector. We evaluate three ratios: $R_{\text{approach}}\!=\!(1.0,0.5,0.05)$ (position-heavy), $R_{\text{pregrasp}}\!=\!(0.05,0.1,1.0)$ (gripper-heavy), and $R_{\text{carry}}\!=\!(0.5,0.5,0.1)$ (position+rotation, low gripper). The perturbation is inserted either \emph{in-chain}, before the remaining denoising steps, or in the \emph{action space}, after denoising completes. We then measure the lag-1 autocorrelation $\hat\rho_{\text{out}}$ and per-group standard deviation $\hat\sigma_{\text{out}}$ of the executed action.\footnote{All measurements average over $256$ Monte-Carlo perturbation samples and $16$ test-set observations.} The SFT baseline rows report the unperturbed reverse flow.

\noindent\textbf{Note on Monte-Carlo fluctuation.}
The SFT baseline rows should be the same across the $\sigma\!=\!0.2$ and $\sigma\!=\!0.5$ runs because they involve no injected perturbation. The small numerical differences in the tables (e.g., $\hat\rho_{\text{SFT,pos}}\!=\!0.7840$ at $\sigma\!=\!0.2$ vs.\ $0.7799$ at $\sigma\!=\!0.5$) come from independent Monte-Carlo estimates. With $256$ perturbation samples and $16$ observations, the standard error is larger than these small shifts, so they do not affect the conclusions.

\noindent\textbf{Temporal correlation is diluted in-chain.}
Table~\ref{tab:dilution}(a) reports $\hat\rho_{\text{out}}$ as $\rho_{\text{inj}}$ sweeps from $0$ to $0.95$. Under in-chain injection, the output correlation stays close to the SFT baseline $\rho_{\text{SFT}}\!\approx\!(0.78,0.72,0.57)$ for all injected values, changing by less than $0.02$ in each group. Under action-space injection, the output correlation changes with the requested value; for example, position correlation increases from $0.11$ to $0.70$ as $\rho_{\text{inj}}$ increases from $0$ to $0.95$. This shows that temporal structure can be specified directly in the action space, while the remaining denoising steps largely absorb the same structure when it is injected in-chain.

\noindent\textbf{Per-group magnitude follows the same pattern.}
Table~\ref{tab:dilution}(b) repeats the comparison for group-wise magnitude profiles. Action-space injection follows the requested group scale: the position-heavy profile increases $\hat\sigma_{\text{pos}}$ from $0.105$ to $0.227$, while the gripper-heavy profile increases $\hat\sigma_{\text{grip}}$ from $0.206$ to $0.314$. In-chain injection changes the same statistics only slightly. The in-chain output is still anisotropic, but the anisotropy mostly reflects the SFT model rather than the injected profile.

\noindent\textbf{Robustness across base noise levels.}
Table~\ref{tab:dilution-sigma05} repeats the protocol at $\sigma\!=\!0.5$, which is $2.5\times$ larger than the setting in Table~\ref{tab:dilution}. The qualitative pattern remains the same. Action-space injection still tracks both the requested temporal correlation and the requested group scale. In-chain injection changes the executed action statistics only mildly; for example, the position-heavy profile increases $\hat\sigma_{\text{pos}}$ from $0.105$ to $0.116$, far below the action-space value of $0.511$. This confirms that structured noise dilution is not caused only by a small base noise level.

\section{Log-Probability of Structured Action-Space Noise}
\label{app:logprob}

This appendix derives Eq.~\ref{eq:logprob}. The sampling process first draws an AR(1) noise sequence for each action dimension $d$:
\begin{equation}
\varepsilon_{1,d} \sim \mathcal{N}(0, 1), \quad \varepsilon_{c+1,d} \;=\; \rho\,\varepsilon_{c,d} \;+\; \sqrt{1-\rho^{2}}\,\zeta_{c,d}, \quad \zeta_{c,d}\!\sim\!\mathcal{N}(0,1),\quad c=1,\ldots,C-1.
\end{equation}
The action-space perturbation is obtained by applying the learned diagonal scale:
\begin{equation}
\Delta_{0,c,d}=\sigma_{c,d}\varepsilon_{c,d}.
\end{equation}
The density of the AR(1) sequence factorizes as
\begin{equation}
\log p(\varepsilon)
=\sum_{d=1}^{D}\left[
\log\mathcal{N}(\varepsilon_{1,d};0,1)
+\sum_{c=1}^{C-1}
\log\mathcal{N}(\varepsilon_{c+1,d};\rho\varepsilon_{c,d},1-\rho^2)
\right].
\end{equation}
During last-step replay, the stored action and the replayed clean action define the perturbation $\Delta_{0}^{\theta}$. Under the current scale, its corresponding AR(1) coordinate is
\begin{equation}
\varepsilon_{c,d}^{\theta}
=\frac{\Delta_{0,c,d}^{\theta}}{\sigma_{c,d}}.
\end{equation}
The transformation from $\varepsilon$ to $\Delta_0$ is diagonal, with Jacobian determinant $\prod_{c,d}\sigma_{c,d}$. The change of variables gives
\begin{equation}
\log p_{\phi}(\Delta_{0}^{\theta}\mid h_{\theta})
=\log p(\varepsilon^{\theta})-\sum_{c,d}\log\sigma_{c,d},
\end{equation}
which yields Eq.~\ref{eq:logprob}. This derivation only describes likelihood evaluation for the replayed perturbation. Sampling still follows the forward construction $\varepsilon\!\to\!\Delta_0$ in Eq.~\ref{eq:structured-noise}.

\section{Regularizer for the Structured Noise Module}
\label{app:regularizer}

This appendix details the regularizer for the structured noise module in Sec.~\ref{sec:structure}. The learned scale $\sigma$ controls the magnitude of action-space exploration. To keep it in a useful range during RL, we use three lightweight penalties:
\begin{equation}
\mathcal{L}_{\text{reg}}(\sigma)
\;=\;
\lambda_{\text{ent}}\big(\log D - \mathcal{H}(\bar{\sigma})\big)
\;+\;
\lambda_{\text{b}}\big(\|\sigma\|_{F} - \sigma^{\text{base}}\sqrt{CD}\big)^{2}
\;+\;
\lambda_{\text{s}}\sum_{c=1}^{C-1}\|\sigma_{c+1}-\sigma_{c}\|^{2},
\end{equation}
where $\bar{\sigma}_{c,d}=\sigma_{c,d}/\sum_{d'}\sigma_{c,d'}$ and $\mathcal{H}(\bar{\sigma})=-\sum_d\bar{\sigma}_{c,d}\log\bar{\sigma}_{c,d}$, averaged over batch and chunk positions. The dimension-entropy term discourages scale collapse onto a few dimensions, the magnitude-anchor term keeps the total exploration scale near the target level, and the smoothness term avoids abrupt scale changes along the chunk.

\section{Real-World Experimental Details}
\label{app:realworld}

This appendix provides additional details for the real-world experiments in Sec.~\ref{sec:realrobot}.

\noindent\textbf{Scene setup.}
Experiments are conducted on a Franka Research~3 (FR3) robot with a ROS~2 Cartesian-impedance control stack. The policy observes two $256\!\times\!256$ RGB views and a 19-dimensional proprioceptive state containing the relative TCP pose and velocity, gripper position, force, and torque. In \emph{pick-banana}, the robot must grasp a banana and place it in a basket. The policy uses $C=10$ and a 7-dimensional action at each step, consisting of 3D translation, a 3D rotation vector, and gripper control. In \emph{plug-charger-in}, the connector is held by the gripper and must be aligned with and inserted into a socket. The policy uses $C=5$ and a 6-dimensional end-effector action, while the gripper is held fixed. Both tasks use delta end-effector pose actions in the current end-effector frame and execute primitive commands at 10~Hz.

\noindent\textbf{Algorithm.}
We instantiate StructRL with $\pi_{0.5}$ and an asynchronous off-policy AWAC learner. The implementation reuses replay-buffer and target-network infrastructure from a SAC worker, but optimizes the AWAC objective rather than a maximum-entropy SAC objective. As in Sec.~\ref{sec:realrobot}, we set the ODE denoising horizon to $T=1$. The rollout policy first predicts a clean action chunk and then applies StructRL exploration noise in normalized action space. For \emph{plug-charger-in}, we use fixed AR(1) parameters $\sigma=0.1$ and $\rho=0.8$ for both translation and rotation. Because $T=1$, the stored latent is the policy-independent initial Gaussian noise $x_1$. The current policy can therefore recompute the clean action from the same $x_1$ and evaluate previously executed policy or human-intervention actions under its current action-space distribution.

Every full action chunk forms one transition $\mathcal{T}_t$ as defined in Eq.~\ref{eq:real_transition}. Executing $a_t^b$ from $o_t$ advances the robot by $C$ primitive steps and produces the next observation $o_{t+C}$. All transitions enter the online buffer. Transitions containing SpaceMouse corrections and transitions from the ten fastest successful episodes are additionally retained in the demo buffer; when both buffers are available, each minibatch contains equal numbers of online and demo transitions. Human-intervention actions are mapped to the same normalized model space as policy actions before forming $a_t^b$. If an intervention begins within a chunk, the remaining policy tail is discarded and replaced by continued human input or a zero-delta hold, avoiding actions predicted for a counterfactual state sequence.

The critic consumes the cached 2048-dimensional pooled PaliGemma feature, the raw 19-dimensional proprioceptive state, and the full executed action chunk $a_t^b$. A shared state trunk feeds one value branch and two action-conditioned $Q$ branches, while a GRU encodes the temporal order within the action chunk. Let $R_t$ and $d_t$ denote the chunk return and terminal indicator computed from $\mathbf{r}_t$ and $\mathbf{d}_t$, respectively. The target for each $Q_i$ is
\begin{equation}
y_t=R_t+(1-d_t)\gamma^C V'(o_{t+C}).
\end{equation}
The target value of the replay action is
\begin{equation}
q_t=\min_{i\in\{1,2\}}Q_i'(o_t,a_t^b).
\end{equation}
Here, primes denote target networks. The twin $Q$ functions are trained by squared regression to $y_t$, and $V$ is fit to the $0.7$ expectile of $q_t$. The replay advantage is
\begin{equation}
A_t=q_t-V(o_t).
\end{equation}
The corresponding AWAC weight is
\begin{equation}
w_t=\exp\!\left(\operatorname{clip}\!\left(\frac{A_t}{\beta},-3,3\right)\right),
\qquad \beta=0.5.
\end{equation}
Finally, the actor is optimized using the same objective as in Eqs.~\ref{eq:real_awac}--\ref{eq:real_awac_weight}:
\begin{equation}
\mathcal{L}_{\mathrm{AWAC}}(\theta)
=-\mathbb{E}_{\mathcal{T}_t\sim\mathcal{D}}
\left[w_t\log\pi_\theta(a_t^b\mid o_t,x_1)\right].
\end{equation}
Successful policy actions and useful human corrections therefore receive larger likelihood updates. Rollout runs on one GPU, while four learner GPUs update the critic and action expert asynchronously. Cached VLM features allow multiple critic updates per physical transition without repeatedly evaluating the visual-language encoder.

\noindent\textbf{Training setup.}
Both tasks use sparse success feedback, with no dense distance shaping or learned reward model. Each task starts from a task-specific $\pi_{0.5}$ checkpoint obtained by SFT on ten pre-collected demonstrations. The resulting policies exhibit only a coarse task-directed motion trend and achieve 0\% success over 25 independent trials before online RL. The critic is warmed up on the initially collected replay before actor updates begin. Physical interaction, replay ingestion, critic learning, actor learning, and policy synchronization then proceed asynchronously.

\noindent\textbf{Control and safety.}
Normalized delta actions are first clipped to the environment bounds, transformed from the current end-effector frame to the robot base frame, scaled to metric delta poses, and integrated from the latest measured pose. The resulting absolute equilibrium pose is then clipped to a task-specific workspace and orientation window before being sent to the Cartesian-impedance controller. Additional safeguards include SpaceMouse override, removal of an interrupted chunk's stale policy tail, controller target clipping, and the robot's collision thresholds. For \emph{plug-charger-in}, success is detected when every Cartesian position coordinate is within 5~mm of the target; the configured orientation threshold is not used by the current success checker. Safety-stop and timeout transitions remain in the online buffer as zero-reward evidence, but enter the demo buffer only when they contain a human intervention.

\section{Long-Horizon Chaining on CALVIN}
\label{app:calvin}

This appendix presents the CALVIN long-horizon evaluation referenced in Sec.~\ref{sec:experiments}. CALVIN evaluates long-horizon behavior by chaining up to five language instructions in one episode. This setting stresses the temporal axis of StructRL because jitter or unstable exploration in an early subtask can affect later subtasks. Table~\ref{tab:calvin} reports per-length success on $\pi_{0.5}$. StructRL achieves the highest average sequence length and is the only method above $0.93$ at Len4. The gap over the time-independent in-chain baselines becomes larger as the sequence length increases, which is consistent with the role of action-space temporal correlation.

\begin{table}[ht]
\centering
\small
\setlength{\tabcolsep}{5pt}
\renewcommand{\arraystretch}{1.05}
\caption{\textbf{Long-horizon chaining on CALVIN.} Per-length success rate (fraction in $[0,1]$) and average sequence length on $\pi_{0.5}$. StructRL achieves the highest average and maintains the strongest performance at longer sequence lengths.}
\label{tab:calvin}
\begin{tabular}{lccccc@{\hskip 10pt}c}
\toprule
\textbf{Model} & \textbf{Len1} & \textbf{Len2} & \textbf{Len3} & \textbf{Len4} & \textbf{Len5} & \textbf{Avg.} \\
\midrule
\rowcolor{graybg}
\boldmath$\pi_{0.5}$\unboldmath{} -- SFT        & 0.927 & 0.843 & 0.767 & 0.688 & 0.613 & 3.838 \\
\quad$+\,\pi_{\text{RL}}$ (Flow-SDE$+$PPO)      & 0.997 & 0.982 & 0.958 & 0.910 & 0.870 & 4.717 \\
\quad$+\,\pi_{\text{RL}}$ (Flow-Noise$+$PPO)    & 0.996 & 0.976 & 0.939 & 0.896 & 0.845 & 4.652 \\
\quad$+\,$Baseline                               & 0.998 & 0.990 & 0.964 & 0.929 & 0.868 & 4.749 \\
\quad$+\,$StructRL                               & \textbf{1.000} & \textbf{0.993} & \textbf{0.968} & \textbf{0.934} & \textbf{0.880} & \textbf{4.775} \\
\bottomrule
\end{tabular}
\end{table}

\section{Training Hyperparameters}
\label{app:hparams}

\begin{table}[ht]
\centering
\small 
\setlength{\tabcolsep}{5pt} 
\caption{\textbf{Training hyperparameters for LIBERO} across the three VLA backbones and four task suites.}
\label{tab:hp_libero}
\resizebox{\columnwidth}{!}{
\begin{tabular}{lcccccccccccc}
\toprule
\multirow{3}{*}{\textbf{Parameters}} & \multicolumn{12}{c}{\textbf{LIBERO}} \\
\cmidrule(lr){2-13} 
 & \multicolumn{4}{c}{GR00T N1.5} & \multicolumn{4}{c}{$\pi_{0}$} & \multicolumn{4}{c}{$\pi_{0.5}$} \\ 
\cmidrule(lr){2-5} \cmidrule(lr){6-9} \cmidrule(lr){10-13} 
 & \textbf{Spatial} & \textbf{Object} & \textbf{Goal} & \textbf{Long} & \textbf{Spatial} & \textbf{Object} & \textbf{Goal} & \textbf{Long} & \textbf{Spatial} & \textbf{Object} & \textbf{Goal} & \textbf{Long} \\ 
\midrule
Train epochs & 200 & 200 & 200 & 200 & 200 & 200 & 400 & 400 & 400 & 400 & 400 & 400 \\
Batch size & 1024 & 1024 & 1024 & 1024 & 2048 & 2048 & 2048 & 2048 & 2048 & 2048 & 2048 & 2048 \\
Update epochs & 1 & 1 & 1 & 1 & 1 & 1 & 4 & 4 & 1 & 1 & 4 & 4 \\
Actor lr & 2e-5 & 2e-5 & 2e-5 & 2e-5 & 2e-5 & 2e-5 & 5e-6 & 5e-6 & 5e-6 & 5e-6 & 5e-6 & 5e-6 \\ 
Value lr & 2e-5 & 2e-5 & 2e-5 & 2e-5 & 2e-5 & 2e-5 & 1e-4 & 1e-4 & 1e-4 & 1e-4 & 1e-4 & 1e-4 \\ 
Noise lr & 2e-5 & 2e-5 & 2e-5 & 2e-5 & 2e-5 & 2e-5 & 1e-4 & 1e-4 & 1e-4 & 1e-4 & 1e-4 & 1e-4 \\ 
\midrule
Interaction steps & 240 & 240 & 320 & 480 & 240 & 240 & 320 & 480 & 240 & 240 & 320 & 480 \\
Parallel environments & 64 & 64 & 64 & 64 & 64 & 64 & 64 & 64 & 64 & 64 & 64 & 64 \\
Rollout epochs & 8 & 8 & 8 & 8 & 8 & 8 & 8 & 8 & 8 & 8 & 8 & 8 \\ 
\midrule
Action chunk $C$ & 5 & 5 & 5 & 10 & 5 & 5 & 5 & 10 & 5 & 5 & 5 & 10 \\
Denoise steps & 4 & 4 & 4 & 4 & 4 & 4 & 4 & 4 & 3 & 5 & 5 & 5 \\
Noise level $\sigma$ & 0.2 & 0.2 & 0.2 & 0.2 & 0.2 & 0.1 & 0.1 & 0.1 & 0.1 & 0.05 & 0.05 & 0.05 \\
\multirow{2}{*}{Structured noise} & \multicolumn{12}{c}{AR(1) $\rho$ (pos, rot, grip): $(0.8, 0.6, 0.7)$} \\
 & \multicolumn{12}{c}{Clip scale (pos/rot/grip): $(0.1, 2.0)/(0.1, 1.5)/(0.1, 3.0)$} \\
\bottomrule
\end{tabular}
}
\end{table}

\begin{table}[ht]
\centering
\small 
\setlength{\tabcolsep}{5pt} 
\caption{\textbf{Training hyperparameters for CALVIN and ManiSkill multitask settings.}}
\label{tab:hp_calvin_maniskill}
\begin{tabular}{@{}l*{3}{>{\centering\arraybackslash}p{0.18\columnwidth}}@{}}
\toprule
\multirow{2}{*}{\textbf{Parameters}} & \multicolumn{1}{c}{\textbf{CALVIN}} & \multicolumn{2}{c}{\textbf{ManiSkill}} \\
\cmidrule(lr){2-2} \cmidrule(lr){3-4}
 & $\pi_{0.5}$ & $\pi_{0}$ & $\pi_{0.5}$ \\ 
\midrule
Train epochs & 300 & 400 & 400 \\
Batch size & 2048 & 5120 & 5120 \\
Update epochs & 4 & 4 & 5 \\
Actor lr & 5e-6 & 2e-5 & 2e-5 \\ 
Value lr & 1e-4 & 2e-5 & 2e-5 \\ 
Noise lr & 1e-4 & 2e-5 & 2e-5 \\ 
\midrule
Interaction steps & 480 & 80 & 80 \\
Parallel environments & 64 & 320 & 320 \\
Rollout epochs & 8 & 1 & 1 \\ 
\midrule
Action chunk $H$ & 5 & 5 & 5 \\
Denoise steps & 5 & 4 & 4 \\
Noise level $\sigma$ & 0.2 & 0.2 & 0.2 \\
\multirow{2}{*}{Structured noise} & \multicolumn{3}{c}{AR(1) $\rho$: $(0.8, 0.6, 0.7)$} \\
 & \multicolumn{3}{c}{Clip scale $s$: $(0.1, 2.0)/(0.1, 1.5)/(0.1, 3.0)$} \\
\bottomrule
\end{tabular}
\end{table}

\end{document}